\documentclass[letterpaper,twocolumn,10pt]{article}
\usepackage{usenix-2020-09}

\usepackage{tikz}
\usepackage{amsmath}
\usepackage{enumitem}
\usepackage{stfloats}
\usepackage{placeins} 

\usepackage{filecontents}

\date{}

\title{\Large \bf GUIDE: A \underline{G}lobal \underline{U}nified \underline{I}nference Engine for \underline{D}eploying Large Language Models in Heterogeneous \underline{E}nvironments}

\author{
    {\rm Yanyu Chen\textsuperscript{1}$^{\#}$, Ganhong Huang\textsuperscript{2}$^{\#}$} \\
    {\small $^{\#}$These authors contributed equally to this work,} \\
    {\textsuperscript{1}School of Systems Science and Engineering, Sun Yat-sen University,}\\
    {\textsuperscript{2}School of Computer Science and Engineering, Sun Yat-sen University}\\
}

\begin{document}
\maketitle

\begin{abstract}
Efficiently deploying large language models (LLMs) in real-world scenarios remains a critical challenge, primarily due to hardware heterogeneity, inference framework limitations, and workload complexities. These challenges often lead to inefficiencies in memory utilization, latency, and throughput, hindering the effective deployment of LLMs, especially for non-experts. Through extensive experiments, we identify key performance bottlenecks, including sudden drops in memory utilization, latency fluctuations with varying batch sizes, and inefficiencies in multi-GPU configurations. These insights reveal a vast optimization space, shaped by the intricate interplay of hardware, frameworks, and workload parameters. This underscores the need for a systematic approach to optimize LLM inference, motivating the design of our framework, GUIDE. GUIDE leverages dynamic modeling and simulation-based optimization to address these issues, achieving prediction errors between 9.9\% and 42.3\% for key metrics such as batch latency, TTFT, and decode throughput. By effectively bridging the gap between theoretical performance and practical deployment, our framework empowers practitioners—particularly non-specialists—to make data-driven decisions and unlock the full potential of LLMs in heterogeneous environments cheaply.
\end{abstract}  
\section{Introduction}

The deployment of Large Language Models (LLMs) has become a pressing challenge, driven by their transformative breakthroughs in natural language processing, computer vision, and multimodal tasks. Models such as GPT~\cite{brown2020language}, OPT~\cite{zhang2022opt}, LLaMA~\cite{touvron2023llama}, and Qwen~\cite{yang2024qwen2}, equipped with billions or even trillions of parameters, demonstrate unparalleled capabilities in generating semantically coherent, contextually rich content and solving complex tasks. These advancements have enabled widespread adoption across diverse applications, including chatbots, content creation, and scientific research.

\begin{figure}[htbp]
    \centering
    \includegraphics[width=0.47\textwidth]{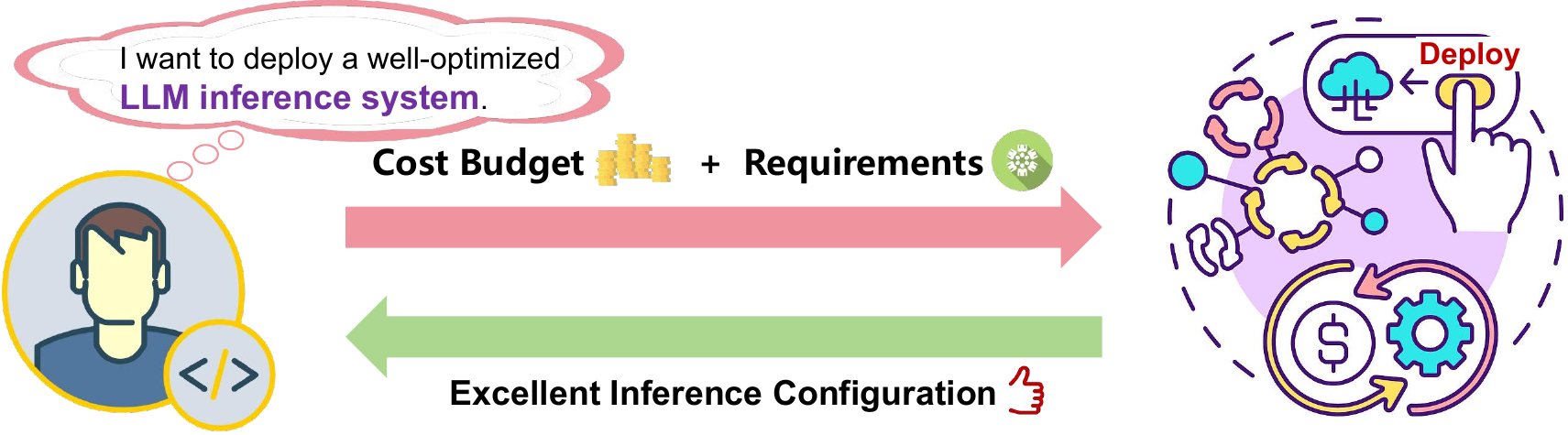} 
    \caption{Workflow of the GUIDE system, which helps users input cost and requirements to generate an optimized LLM inference configuration.}
    \label{fig:workflow_overview}
\end{figure}

However, efficiently deploying such LLMs in real-world scenarios remains a critical bottleneck due to the complexity of real-world constraints. For many enterprise users and individual practitioners, the deployment process is particularly challenging because it requires deep expertise in hardware configurations, inference frameworks, and optimization strategies. Without this expertise, they often struggle to fully leverage the capabilities of LLMs, leading to underutilized hardware, inefficient resource allocation, and suboptimal performance. This complexity underscores the urgent need for accessible and systematic tools that simplify the deployment process, enabling users with varying levels of expertise to achieve efficient and cost-effective inference.

Inference performance is heavily influenced by the heterogeneity of hardware platforms, inference frameworks, and deployment strategies. These disparities often result in underutilized computational resources, increased operational costs, and degraded user experiences, such as higher latency or reduced throughput. These challenges underscore the urgent need for systematic optimization in LLM inference, including hardware-aware tuning, inference framework enhancements, and intelligent scheduling, to bridge the gap between theoretical state-of-the-art model performance and practical deployment requirements, while improving resource utilization, reducing latency, and lowering deployment costs.

Despite extensive efforts to optimize LLM inference, existing approaches often target isolated aspects, such as hardware acceleration~\cite{dao2022flashattention}, framework-specific tuning, or parallelization strategies~\cite{narayanan2021efficient}.While these methods achieve localized improvements, they fall short in addressing the intricate dependencies and dynamic interactions among model architectures, hardware platforms, inference frameworks, and optimization techniques. For instance, tensor parallelism improves throughput by distributing computations across GPUs, but the accompanying communication overheads often diminish its effectiveness, particularly for long sequence lengths or small batch sizes~\cite{narayanan2021efficient}. Similarly, inference frameworks like vLLM~\cite{kwon2023efficient} and Fastgen~\cite{holmes2024deepspeed} excel in specific scenarios but face difficulties in adapting to variable workloads or heterogeneous hardware configurations, resulting in suboptimal performance under diverse conditions. These fragmented approaches fail to fully exploit the optimization potential, particularly in heterogeneous deployment environments where performance bottlenecks stem from multi-dimensional factors, including batch size, sequence length, memory capacity, and computational throughput.

To better understand these challenges, we conducted extensive experiments across diverse hardware platforms, inference frameworks, deployment configurations, and optimization methodologies. Our findings reveal significant opportunities for optimization, particularly in memory efficiency and latency reduction, but also underscore the inherent complexity arising from the interplay between hardware, software, workload characteristics, and optimization techniques. For instance, we observed abrupt memory utilization drop-offs under specific configurations, significant latency divergence with varying batch sizes, and performance degradation even in modest parallel scenarios using multiple GPUs. These results highlight the critical need for careful selection and integration of hardware platforms, inference frameworks, deployment configurations, and optimization strategies, tailored to the specific requirements of the workload, to maximize inference efficiency. They also point to limitations in existing deployment practices, which often fail to capture the nuanced interactions among these dimensions, resulting in inefficient resource utilization and suboptimal performance. Addressing these limitations will require a more holistic approach that dynamically balances hardware, software, and optimization techniques to adapt to varying workload demands.

While these insights reveal a vast and complex optimization space, they also highlight the inherent limitations of existing optimization practices. Manual tuning of deployment configurations is resource-intensive, prone to human error, and often lacks the flexibility to generalize across diverse scenarios. Furthermore, while existing data provide valuable insights, they are limited in scope, covering only a subset of frameworks, strategies, and optimization techniques. This narrow coverage hinders comprehensive evaluation of their effectiveness and complicates performance prediction in unseen configurations. This incomplete coverage not only constrains systematic exploration of the optimization space but also inhibits practitioners from effectively adapting to the dynamic evolution of models, frameworks, and hardware.

In response to these challenges, we propose GUIDE, a comprehensive modeling and simulation framework to systematically explore and optimize the inference process of LLMs. By constructing a performance model that incorporates key factors such as hardware (e.g., GPUs), inference frameworks, deployment strategies, optimization techniques, and workload-specific parameters (e.g., batch size, input length), the framework systematically searches for configurations that deliver exceptional performance. Designed to address specific requirements and constraints, the framework enables researchers and practitioners to explore a vast optimization space, predict the performance of untested configurations, and make informed deployment decisions with confidence. By abstracting the optimization process into a modeling and search problem, the framework significantly reduces the time and effort traditionally required for deployment tuning, while ensuring scalability and adaptability across diverse hardware platforms and real-world scenarios.

\textbf{Contributions.} This work makes the following key contributions:
\begin{itemize}[itemsep=0em, parsep=0em]
    \item We conduct systematic experiments that reveal key performance issues in LLM inference, including memory utilization drop-offs caused by framework-hardware mismatches and latency divergence under varying batch sizes and parallel configurations.
    \item We propose GUIDE, a novel deployment system, that integrates hardware modeling, model analysis, inference frameworks, deployment strategies, and optimization techniques to efficiently explore multi-dimensional optimization spaces.
    \item We validate the proposed simulator's effectiveness, achieving an error range of 9.9\% to 42.3\% across performance metrics, effectively supporting deployment decisions in diverse configurations.
\end{itemize}   
\section{Background \& Motivation}
\label{sec:bg}

\subsection{Transformer Models}
\begin{figure}[htbp]
    \centering
    \includegraphics[width=0.45\textwidth]{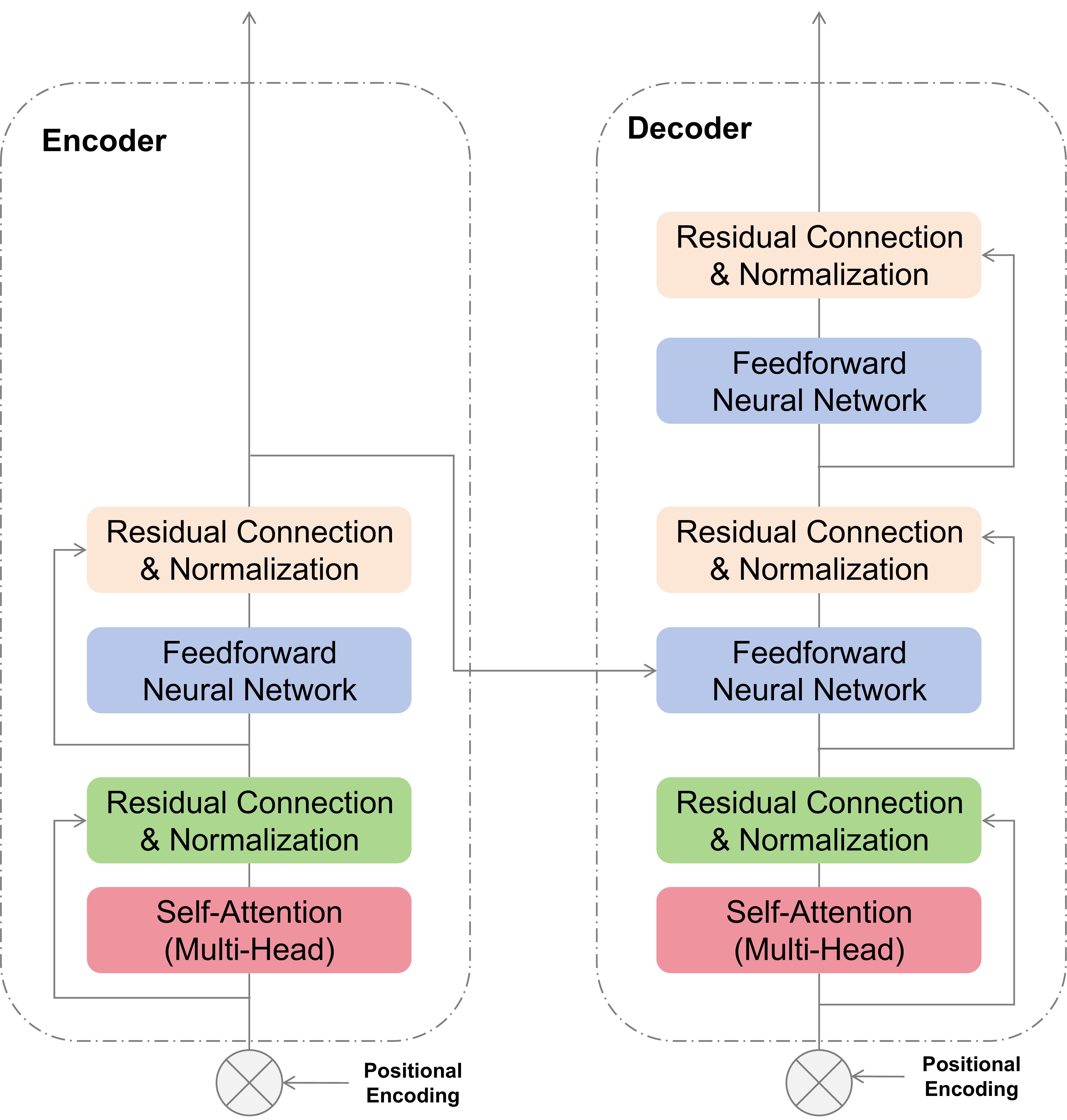} 
    \caption{Basic transformer architecture.}
    \label{fig:transformer}
\end{figure}

Transformer models~\cite{vaswani2017attention} have significantly advanced artificial intelligence by enabling breakthroughs in natural language processing (NLP), computer vision, and multimodal tasks. As shown in Figure~\ref{fig:transformer}, the architecture consists of stacked Transformer blocks, each composed of Multi-Head Attention (MHA), Feed-Forward Networks (FFN), and Normalization (Norm) layers. Residual connections facilitate gradient flow during training, while positional encodings provide sequence order information that the architecture itself lacks.

The Transformer architecture is divided into an encoder and a decoder. The encoder processes input sequences and builds contextual representations by capturing global dependencies through self-attention. The decoder generates output sequences by attending to both the encoder's output and its own previously generated tokens. This design supports a wide range of sequence-to-sequence tasks, such as machine translation and text generation.

\subsection{LLM Inference Challenges}

Deploying Large Language Models (LLMs) presents several technical challenges, primarily arising from memory bottlenecks, computational complexity, and latency requirements.

A critical issue is memory usage during inference, dominated by the storage of Key-Value (KV) pairs in self-attention mechanisms. These KV pairs are retained across all layers, with memory consumption scaling linearly with sequence length and model depth. For long input sequences or deep models, this can quickly exceed hardware capacities, especially on GPUs with limited memory. Techniques like memory-efficient attention and model quantization have been proposed to alleviate this constraint, but they often involve trade-offs in precision or computational overhead.

Another major challenge is the quadratic computational complexity of the self-attention mechanism with respect to sequence length. This limits throughput, particularly for tasks requiring long context windows or high concurrency. Underutilization of hardware resources is common in such scenarios, further exacerbated by mismatches between model architectures and hardware capabilities.

Latency requirements add another layer of complexity, especially for real-time applications. Factors such as batch size, sequence length, and parallelism strategies heavily influence latencies. High-concurrency settings, small batch sizes, or workloads with variable input lengths often result in significant performance degradation due to suboptimal scheduling or increased communication overheads in distributed systems.

The heterogeneity of hardware platforms and inference frameworks further complicates deployment. Different hardware and software systems often exhibit inconsistencies in their ability to handle diverse workloads, requiring careful tuning to achieve optimal performance under practical scenarios. These challenges necessitate systematic optimization strategies to balance memory usage, throughput, and latency across heterogeneous environments.

\subsection{LLM Inference Optimization}

Optimizing the inference of LLMs requires addressing challenges such as memory bottlenecks, computational complexity, and latency constraints. Over the years, researchers have developed a variety of techniques, including algorithmic optimizations, deployment strategies, and specialized inference frameworks. This section provides an overview of these methods and their relevance to LLM inference.

\subsubsection{Algorithmic Optimizations}

Algorithmic techniques are fundamental to overcoming memory and computational challenges in LLM inference. Quantization methods, such as SmoothQuant~\cite{xiao2023smoothquant} and LLM.int8~\cite{dettmers2022gpt3}, reduce the precision of weights and activations, significantly lowering memory consumption while maintaining acceptable accuracy. Pruning approaches, such as SparseGPT~\cite{frantar2023sparsegpt}, identify and remove redundant parameters, thereby reducing computational complexity and accelerating inference. FlashAttention optimizes the self-attention mechanism by minimizing memory access overheads, enabling efficient processing of long sequences. These techniques have demonstrated their effectiveness in addressing specific bottlenecks, making them widely adopted in real-world deployments.

\begin{figure}[htbp]
    \centering
    \includegraphics[width=0.47\textwidth]{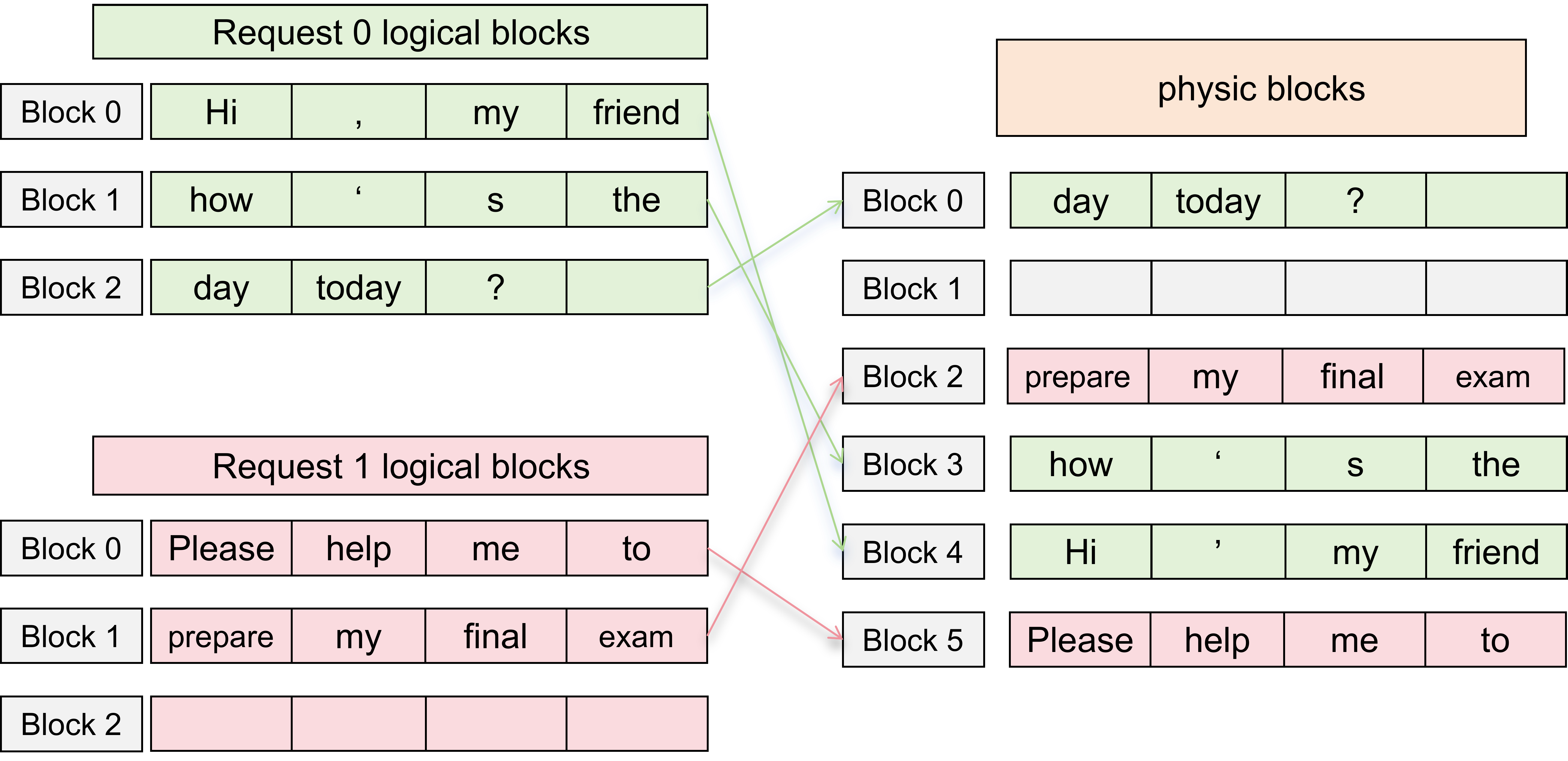} 
    \caption{Logical-to-physical block mapping in vLLM.}
    \label{fig:vllm}
\end{figure}

\subsubsection{Inference Frameworks}

\begin{figure}[htbp]
    \centering
    \includegraphics[width=0.47\textwidth]{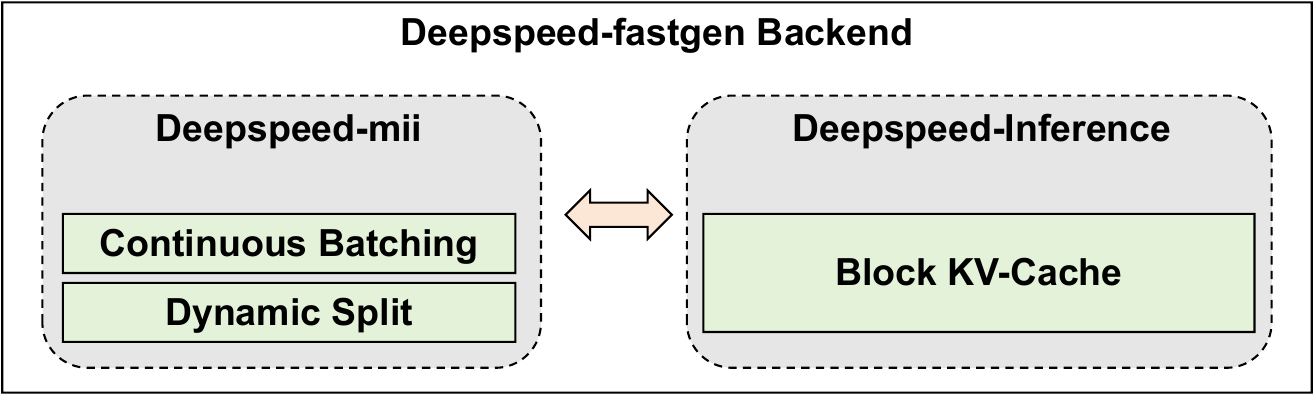} 
    \caption{The architecture of the DeepSpeed-FastGen backend, showing continuous batching and dynamic splitting in DeepSpeed-MII, and block KV-cache in DeepSpeed-Inference.}
    \label{fig:fastgen}
\end{figure}

The growing demand for efficient LLM deployments has led to the emergence of numerous inference frameworks, each designed to address different aspects of the deployment process. Popular frameworks such as vLLM, DeepSpeed-FastGen~\cite{holmes2401deepspeed}, TGI (Text Generation Inference) ~\cite{tgi2023}, FasterTransformer~\cite{fastertransformer2023}, and LLaMA.cpp~\cite{llama.cpp2023} represent diverse strategies for optimizing latency, throughput, and resource utilization. These frameworks provide tailored solutions for various deployment scenarios, ranging from high-performance cloud systems to resource-constrained edge environments.

vLLM introduces dynamic batching and parallelized token generation to improve computational efficiency and reduce latency. As shown in Figure~\ref{fig:vllm}, the figure illustrates the logical-to-physical block mapping framework used in vLLM. FastGen leverages scalability through dynamic splitting and fusion techniques integrated into its backend architecture. As illustrated in Figure~\ref{fig:fastgen}, the DeepSpeed-FastGen backend consists of two main components: DeepSpeed-MII, which supports continuous batching and dynamic splitting, and DeepSpeed-Inference, which utilizes block KV-cache for efficient inference. TGI supports distributed inference with model sharding, enabling large-scale deployments across multi-GPU or multi-node clusters. FasterTransformer focuses on optimizing inference for NVIDIA hardware, leveraging advanced kernel-level optimizations. LLaMA.cpp, in contrast, caters to resource-constrained environments by employing aggressive quantization and memory management techniques.

Although this work primarily evaluates vLLM and FastGen due to their unique optimization strategies, the methodologies discussed are broadly applicable to other frameworks, reflecting the diversity of approaches in this area.

\subsubsection{Deployment Strategies}

Deployment strategies, such as tensor parallelism, data parallelism, and pipeline parallelism, are critical for distributing computations across hardware resources. Tensor parallelism partitions model parameters across GPUs, enabling efficient handling of large-scale models. However, the approach introduces substantial communication overheads, as intermediate results must be synchronized during inference. Data parallelism, by splitting input data across devices, is well-suited for batch processing but requires careful synchronization to maintain consistency. Pipeline parallelism divides model layers across GPUs, allowing simultaneous execution of different stages of the model but introducing latency due to inter-stage dependencies.

\subsection{Experimental Insights}

\subsubsection{Memory Utilization Drop-Offs}

\begin{figure}[htbp]
    \centering
    \includegraphics[width=0.47\textwidth]{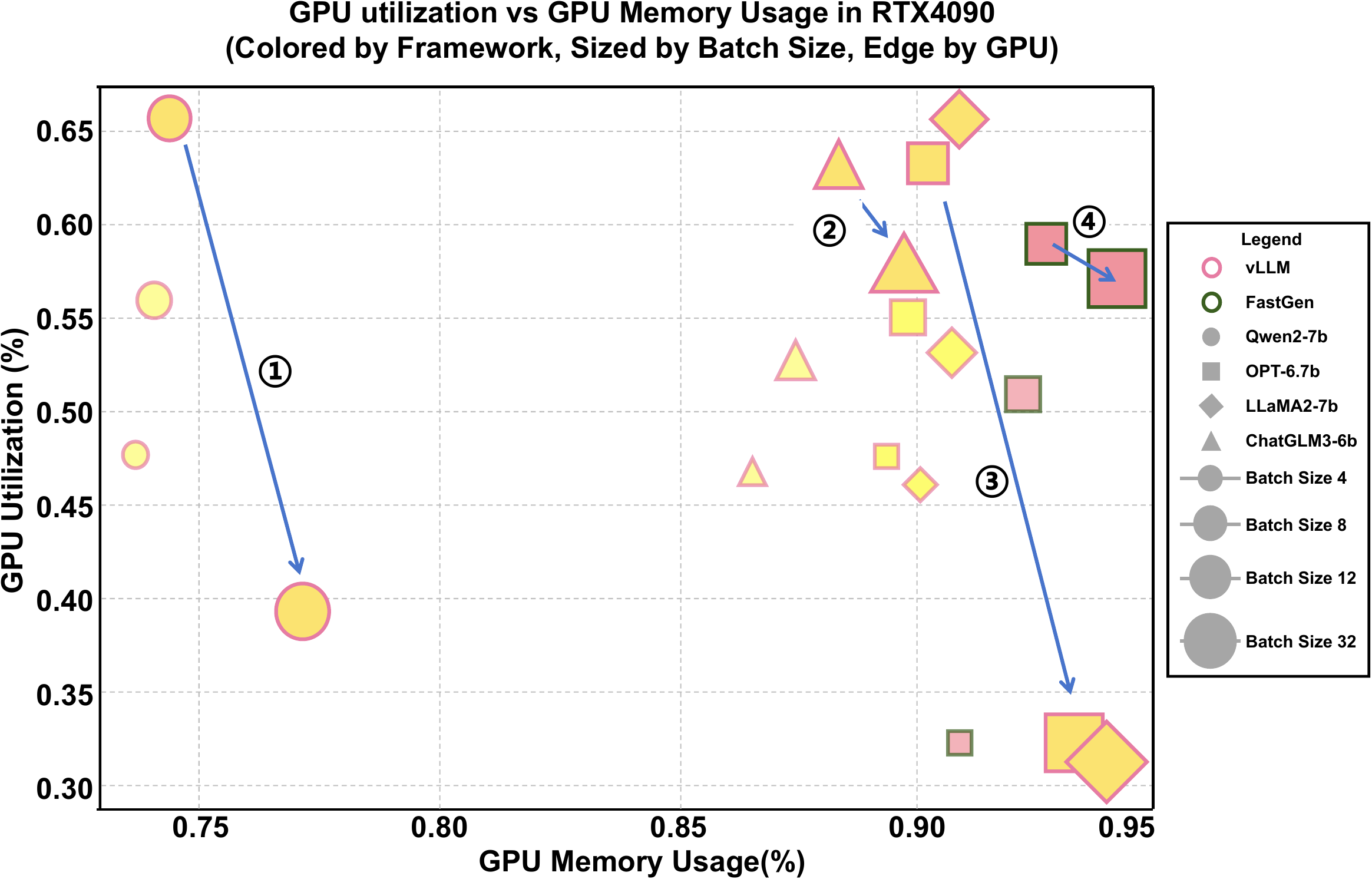} 
    \caption{Memory utilization drop-offs.}
    \label{fig:insight1}
\end{figure}

Both vLLM and FastGen exhibit sharp memory utilization drop-offs at critical points across specific hardware and model combinations. For vLLM, dramatic declines are observed on GPUs such as RTX 4090 and V100, where memory utilization drops abruptly as batch sizes increase from smaller values (e.g., 8) to larger ones (e.g., 32) for models like Qwen, OPT, and LLaMA (Figure~\ref{fig:insight1}. On A6000 GPUs, similar drop-offs occur with LLaMA, while FastGen, despite maintaining relatively higher utilization levels under most conditions, also encounters significant declines, such as with the OPT model on RTX 4090 and the LLaMA model on A6000. A100 GPUs exhibit more stable performance, yet inefficiencies persist when handling complex models like Qwen and OPT.

These abrupt drop-offs stem from mismatches between KV-cache management strategies and hardware memory allocation mechanisms. vLLM’s fixed-granularity KV-cache allocation strategy, while effective for small-batch tasks, leads to severe memory fragmentation and underutilization as batch sizes grow. On GPUs like A100, which allocate memory in large blocks (e.g., 2MB), these inefficiencies are magnified. FastGen, though employing a more adaptive caching mechanism, struggles with certain model and hardware combinations, where suboptimal allocation persists.

Moreover, the heterogeneous characteristics of models, such as KV-cache size and access patterns, further exacerbate these issues. Uniform caching strategies fail to adapt to these variations, particularly as batch sizes increase, leading to sudden memory inefficiencies for both frameworks. These findings reveal not only the performance bottlenecks in memory management but also the significant optimization opportunities that exist for improving GPU resource utilization across diverse workloads and hardware configurations.

\subsubsection{Batch Size and Latency Divergence}

\begin{figure}[htbp]
    \centering
    \includegraphics[width=0.47\textwidth]{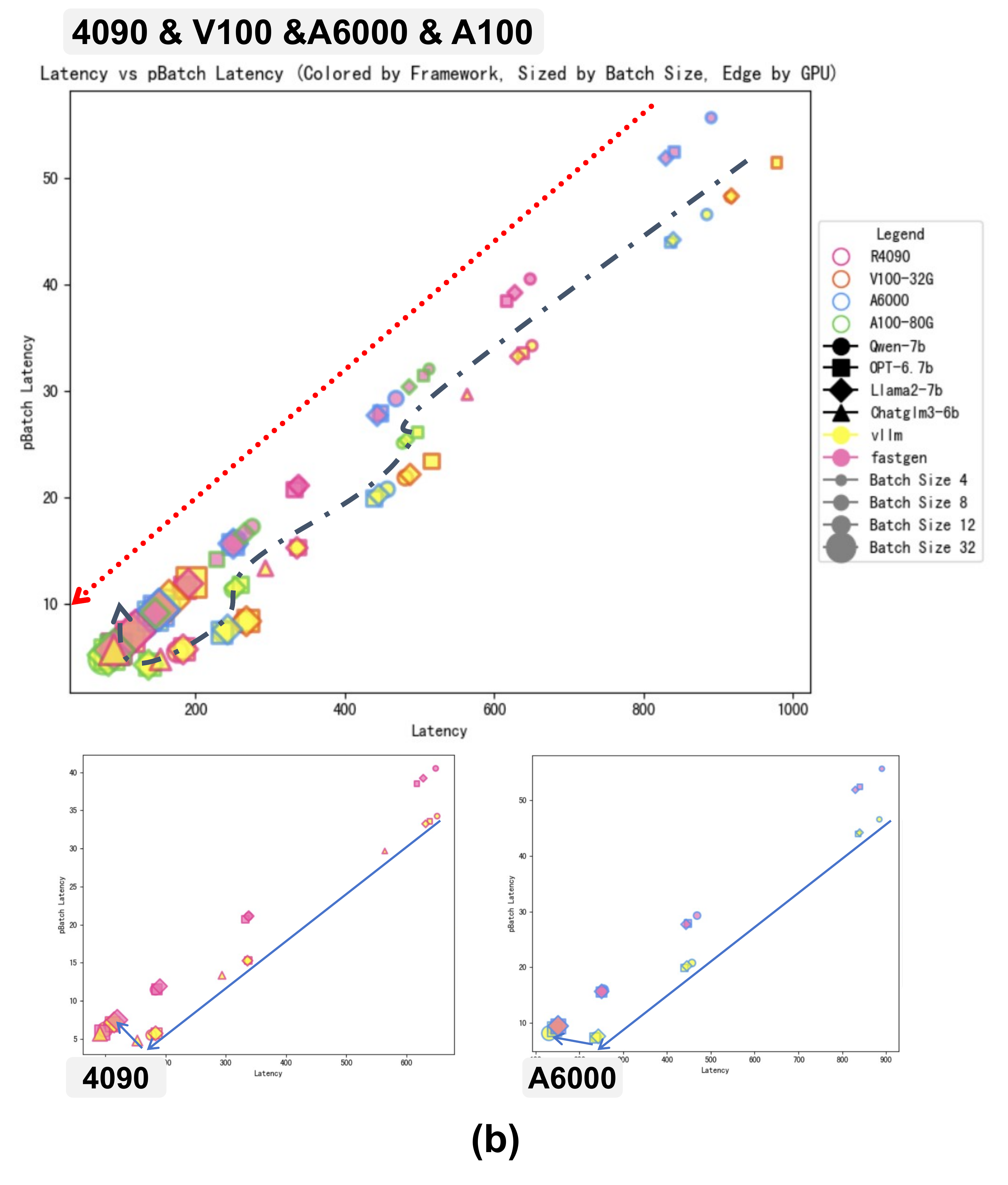} 
    \caption{inference latency turn-off.}
    \label{fig:insight2}
\end{figure}

Latency performance shows a striking divergence between vLLM and FastGen as batch size increases, with a clear inflection point observed. For small-batch tasks (Batch Size $\leq$ 16), vLLM achieves significant latency advantages over FastGen, making it highly suitable for real-time applications. However, as batch size grows, vLLM’s latency gradually deteriorates, and at Batch Size 32, FastGen surpasses vLLM, demonstrating stable and scalable performance. This divergence is particularly pronounced on high-end GPUs such as RTX 4090 and A6000, where vLLM’s latency increases sharply, while FastGen maintains consistent performance over a wider range of batch sizes (Figure~\ref{fig:insight2}).

This divergence can be attributed to the fundamental differences in the resource allocation and task scheduling strategies employed by the two frameworks. vLLM prioritizes prompt processing by allocating substantial resources to KV-cache initialization and storage. While this approach performs well for small-batch tasks, it leads to significant resource contention as batch sizes grow. Specifically, vLLM’s static KV-cache allocation strategy fails to scale with increasing workload complexity, causing delays in token generation and sharp latency spikes. In contrast, FastGen employs a dynamic splitting and fusion strategy that optimizes matrix operations by breaking them into smaller, parallelizable chunks. This approach minimizes contention, reuses intermediate results, and ensures balanced resource utilization, enabling FastGen to scale efficiently with larger batch sizes.

These findings highlight a significant optimization space for balancing latency performance across batch sizes. The trade-off between real-time performance for small batches and scalability for larger batches reveals opportunities to improve task scheduling and resource allocation strategies, particularly for high-end GPUs and diverse workload requirements.

\subsubsection{Performance Under Parallel Environments}

\begin{figure}[htbp]
    \centering
    \includegraphics[width=0.47\textwidth]{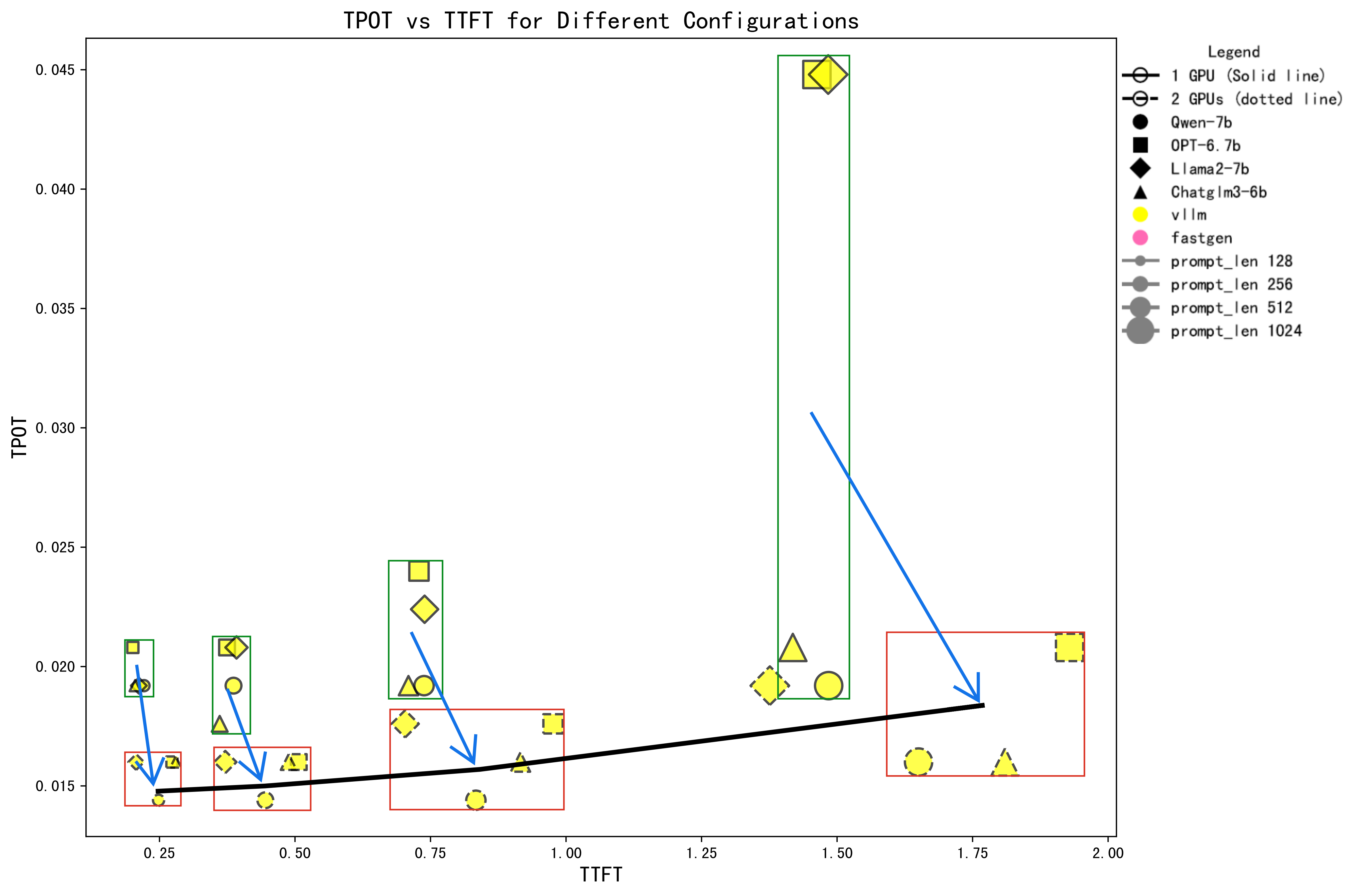} 
    \caption{Latency Performance Under Parallel Environments for vllm.}
    \label{fig:insight3.1}
\end{figure}

\begin{figure}[htbp]
    \centering
    \includegraphics[width=0.47\textwidth]{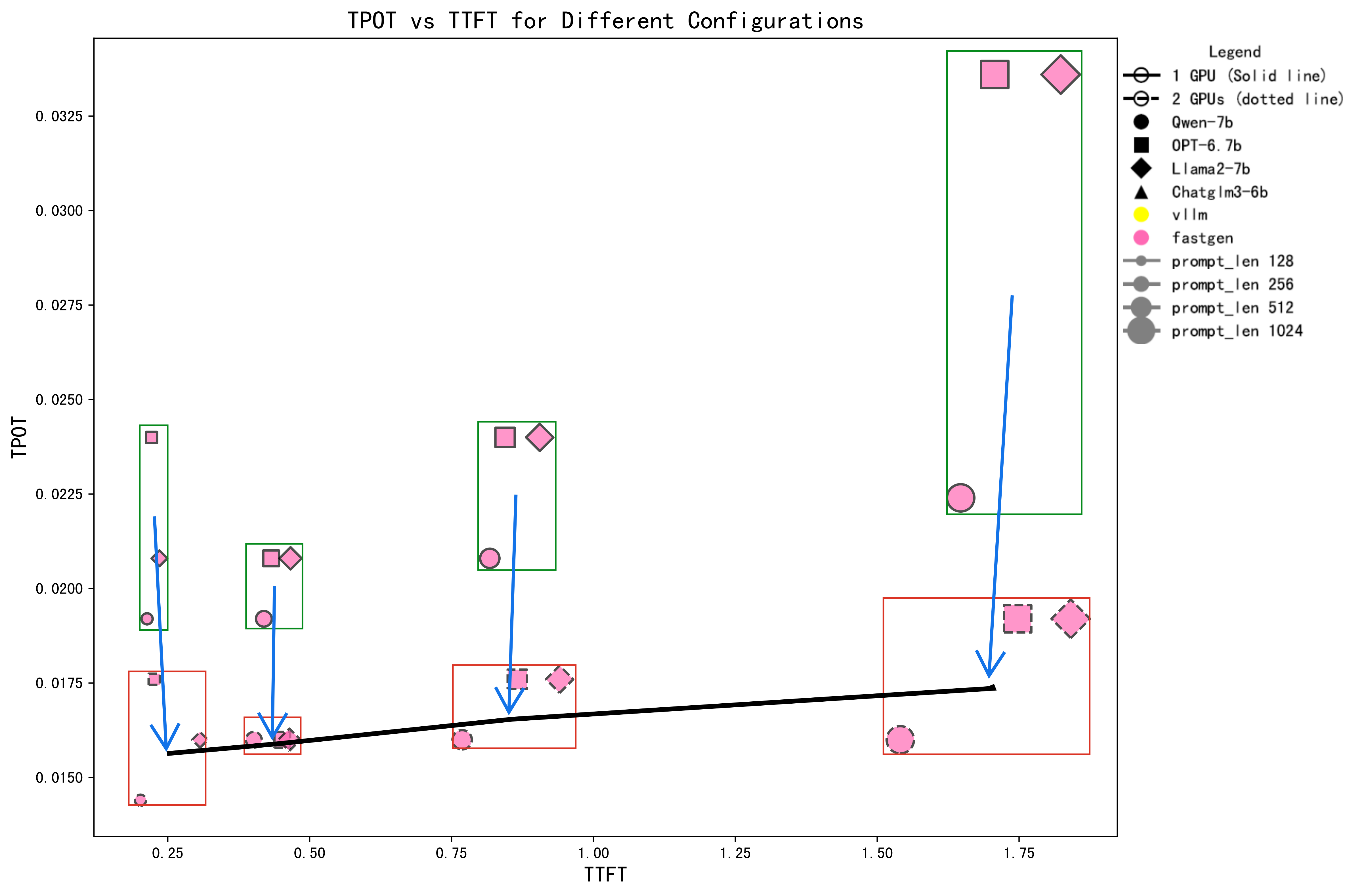} 
    \caption{Latency Performance Under Parallel Environments for deepspeed-fastgen.}
    \label{fig:insight3.2}
\end{figure}

When tensor parallelism (TP) increased from TP=1 to TP=2, a divergence in first token latency (TTFT) was observed. For models like LLaMA, vLLM exhibited noticeable increases in TTFT, while FastGen maintained stable latency and even achieved slight reductions in some cases. As shown in Figure~\ref{fig:insight3.1} and Figure~\ref{fig:insight3.2}, vLLM’s TTFT rises with increasing parallelism, whereas FastGen demonstrates relatively consistent performance. Specifically, in the figures, the bottom-left corner represents the optimal performance, and the top-right corner indicates the worst performance. The green box highlights the performance with a single GPU, while the red line represents the performance with 2 GPUs. The blue arrow illustrates the trend from single GPU to multi-GPU, and the black line shows how this trend changes with increasing prompt length.

The observed divergence is primarily caused by the complex interaction between tensor parallelism (TP) and the computational and communication patterns of different phases. During the Prefill phase, input prompts are processed in a single pass, requiring intensive computation and frequent synchronization of intermediate results across GPUs. As parallelism increases, communication overhead grows significantly, leading to reduced Prefill efficiency. In the Decoding phase, where token generation proceeds step-by-step (each step depending on the previous one), cumulative synchronization delays further exacerbate bottlenecks in the generation process. These results suggest that TTFT divergence is not only influenced by hardware communication and parallel strategies but also by the intricate interaction between model architectures and computational phases.

These findings highlight that in tensor parallel environments, there is considerable room for optimization in the computational and communication patterns of different phases. By analyzing the key bottlenecks in Prefill and Decoding phases and understanding the factors affecting their efficiency, it is possible to explore better parallel strategy designs to improve inference performance.

\subsubsection{Dynamic Batch Size Optimization}

Batch size selection plays a critical role in determining latency performance, as it directly affects how computational resources are utilized. Different tasks exhibit distinct optimal batch sizes due to their unique computational characteristics and resource demands. For tasks with high computational intensity, such as matrix multiplications with large dimensions, the GPU’s computational units can quickly reach saturation even at smaller batch sizes. In contrast, tasks with lower computational demands scale efficiently to larger batch sizes before encountering hardware bottlenecks. Beyond these points, further increasing the batch size often leads to diminishing returns as resources such as compute units and memory bandwidth become fully utilized, limiting scalability and causing latency to increase.

This variation in optimal batch size highlights the intricate interaction between task characteristics and hardware constraints. Computationally intensive tasks saturate GPU resources at smaller batch sizes, leaving little room for further scalability, while less intensive tasks allow for greater scaling before bottlenecks arise. These differences make it clear that a one-size-fits-all batch size configuration is inherently suboptimal, as it fails to account for the varying demands of different workloads.

These observations reveal a significant optimization space for tailoring batch size configurations to specific task characteristics and hardware conditions. By analyzing the relationship between computational intensity, hardware utilization, and latency performance, it is possible to explore strategies that dynamically adjust batch size to better align with workload demands.

\subsection{Motivation}

The insights from our experiments highlight critical bottlenecks in LLM inference, revealing a vast optimization space shaped by the intricate interplay of multiple factors, including hardware configurations, model architectures, inference frameworks, deployment strategies, optimization methods, and workload parameters (e.g., batch size, sequence length). These factors, combined with the computational characteristics of the workloads, create a highly complex and sensitive environment where small changes in one dimension can significantly impact overall performance. Current methods, which often rely on static configurations or isolated tuning, fail to address these dynamic dependencies, leaving substantial optimization potential untapped in real-world deployments.

To address these challenges, we propose a systematic framework for modeling and simulating LLM inference. This framework aims to explore the multi-dimensional optimization space created by interactions among hardware, models, frameworks, and workload characteristics. By identifying the key factors that influence performance and understanding their interdependencies, our approach provides actionable insights for intelligent deployment decisions. It bridges the gap between theoretical advancements and practical requirements, paving the way for scalable, efficient, and deployable LLM inference across diverse configurations and scenarios.
\section{Framework}
\label{sec:framework}

To mitigate the performance bottlenecks identified in Section~\ref{sec:bg}, we propose an intelligent deployment system designed to optimize inference performance for large-scale models. This system systematically models various configurations and simulates the performance across different inference setups, allowing for the identification of the optimal solution under complex inference configurations. It explores the multi-dimensional parameter space of model inference, including hardware platforms, inference frameworks, parallel strategies, and optimization techniques. By integrating advanced modeling approaches with simulation-based optimization, the system effectively addresses challenges arising from complex factor interactions and dynamic bottlenecks, delivering robust and actionable performance improvements.

The intelligent deployment system is built around two core objectives. First, it aims to accurately model and predict the performance of large-scale models under diverse configurations by capturing the nuances of hardware, frameworks, and workload parameters. Second, it seeks to automate the optimization of deployment strategies, enabling users to achieve near-optimal performance with minimal manual intervention. A distinctive feature of this system is its adaptability to workload characteristics, hardware constraints, and framework-specific behaviors, ensuring consistent and reliable results across a wide range of deployment scenarios.

\subsection{Overview}

GUIDE addresses these challenges by automating the exploration of the multi-dimensional parameter space involved in large-scale model inference. It dynamically adapts to different hardware configurations, inference frameworks, and parallel execution strategies, ensuring that it identifies the most efficient deployment configurations for a given task. At its core, GUIDE optimizes both memory and computational resource usage, enabling the deployment of large-scale models with minimal manual effort while achieving maximized performance.

One of the key capabilities of GUIDE lies in its intelligent hybrid parallel simulation. By combining data parallelism (DP) and tensor parallelism (TP), it simulates various GPU configurations to determine the optimal parallel strategy. This simulation process minimizes total execution time in order to find the best parallel config. The system models the computational load and memory overhead of various stages in the inference process, including the prefill and decode phases. It dynamically adjusts batch sizes and sequence lengths based on these models to fit within the constraints of available GPU memory. This adjustment abstracts the influence of inference frameworks such as vllm and fastgen on the inference process, and to model these frameworks, it simulates the dynamic batch processing and dynamic split flow.

\begin{figure*}[htbp]
    \centering
    \includegraphics[width=0.85\textwidth]{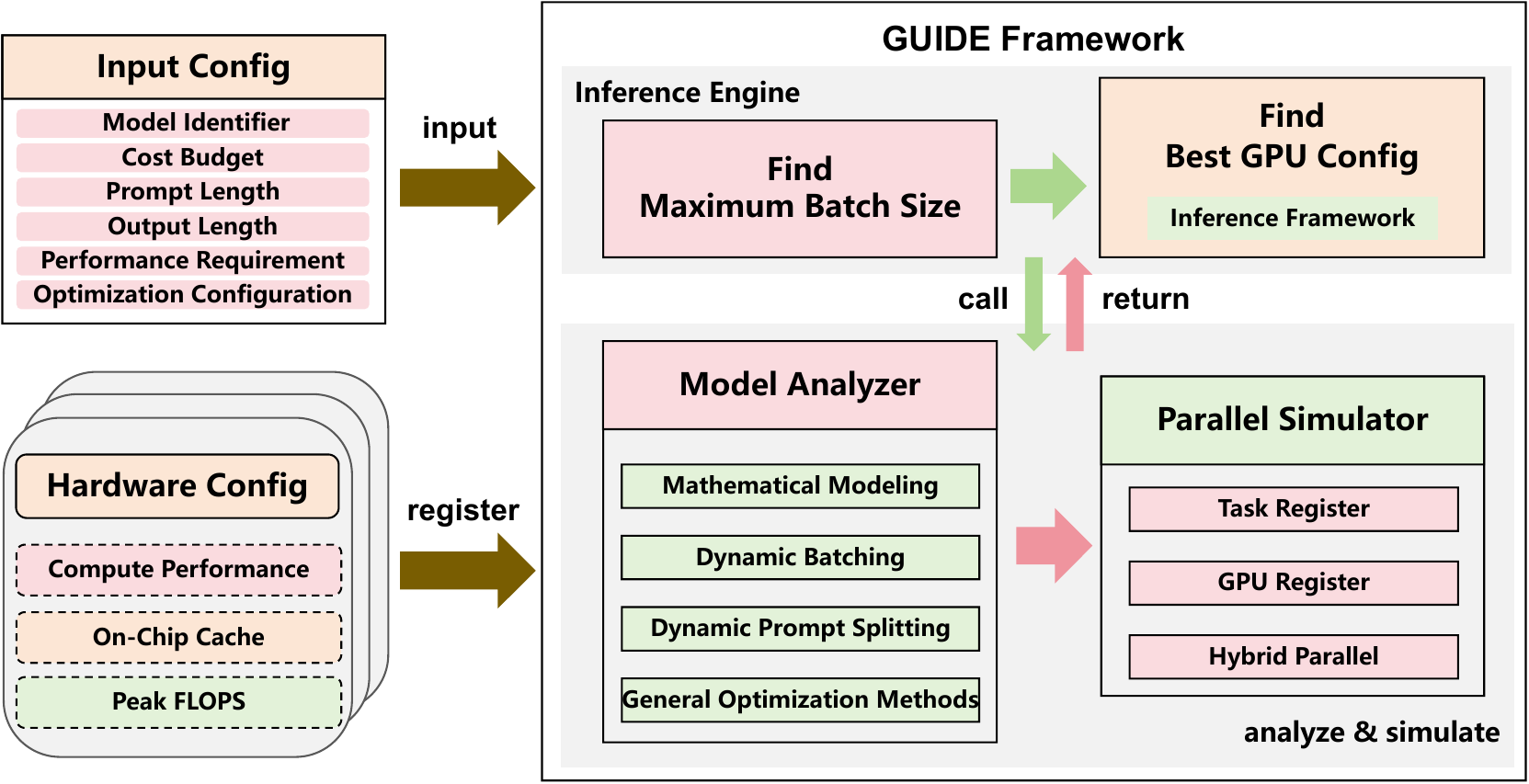} 
    \caption{Overview of GUIDE. The figure shows the architecture of GUIDE, where task-specific input configurations are generated based on user-provided requirements, and hardware specifications are pre-registered into the system. The Inference Engine interacts with the Model Analyzer and Parallel Simulator to determine the maximum batch size and the best GPU configuration. The selection of the best GPU configuration also incorporates the choice of inference frameworks. Ultimately, GUIDE generates optimal configurations tailored to diverse workloads, including parameters such as batch size, GPU selection, inference frameworks, and deployment strategies.}
    \label{fig:framework_overview}
\end{figure*}

To further enhance its predictive capabilities, GUIDE incorporates the Roofline model\cite{williams2009roofline}, similar to the approach used by LLM-Viewer\cite{yuan2024llm}, which models the computational and memory overheads of inference. This integration allows it to analyze task performance and identify bottlenecks in computation and memory bandwidth, providing insights that guide decisions about parallel strategies and resource allocation. Coupled with simulation-based optimization, the system evaluates task execution time and throughput for different parallel configurations. Based on these metrics, it selects the top-performing configurations, ensuring that performance goals are met under the given constraints.

In addition to these analytical capabilities, GUIDE automates the generation of deployment configurations. By analyzing hardware and workload characteristics, it produces a set of potential configurations that can handle diverse deployment scenarios without requiring manual fine-tuning. This automation streamlines the process of adapting to new environments, making the system highly versatile and efficient.

Through these techniques, GUIDE empowers users to achieve near-optimal inference performance for large-scale models. Its adaptability ensures that resource utilization is maximized and task execution is completed in the shortest possible time. Whether deployed in single-GPU setups or large multi-GPU clusters, the system's flexibility and robustness make it suitable for a wide range of deployment environments.

As shown in Figure~\ref{fig:framework_overview}, GUIDE considers various factors and simulates the performance characteristics of the actual inference process, modeling the multi-dimensional parameter space involved in large-scale model inference. It dynamically adapts to different hardware configurations, inference frameworks, and parallel execution strategies, ensuring that it identifies the most efficient deployment configurations for a given task.

\subsection{Inference Engine}

\textbf{Step 1: Memory Usage and Maximum Parallelism Calculation}

To evaluate the memory overhead, which primarily arises from the key-value (kv) storage and model weight storage, we first perform a mathematical analysis using the Model Analyzer component. This step does not require the inference framework itself such as vllm or fastgen since they do not affect the total amount of memory consumption during prefill and decode phases, but takes into account user inputs for optimization such as FlashAttention and H2O which reduce the memory consumption. The analysis focuses on calculating the kv overhead for a batch size of 1. The kv overhead is derived by averaging the memory usage for a single token during both the prefill and decode phases.

The available GPU memory is then calculated as:
\[
\text{Available Memory} = \text{GPU Memory} - \text{Model Slice Memory}
\]
(where Model Slice Memory is determined by TP splitting).

Using this, the maximum parallelism can be derived as:
\[
\text{Maximum Parallelism} = \text{DP} \times \left(\frac{\text{Available Memory}}{\text{kv overhead per request}}\right),
\]

and the maximum allowed batch size (\( \text{Max Batch Size} \)) is:
\[
\text{Max Batch Size} = \frac{\text{Maximum Parallelism}}{\text{Data Parallelism}}.
\]
This process is carried out using the \textit{Parallel Simulator}, which helps identify the optimal parallel configuration (data parallelism, \( dp \), and tensor parallelism, \( tp \)) based on the maximum batch size.

\textbf{Step 2: Task Scheduling and Simulation for Optimal Performance}

Once the maximum batch size has been determined, the next step is to analyze the performance of the chosen configuration using the Model Analyzer. In this step, we analyze both vllm and FastGen inference models under the same configs as step 1. Specifically, the kv overhead for each token during the prefill and decode phases is calculated. The total memory required for processing a request is the sum of the kv overhead and the model weight size.

The Model Analyzer also calculates the computational overhead based on the number of tokens to be generated. This information is then used to generate a task list for GUIDE, which consists of tasks for the prefill and decode stages. Each task includes both memory and computation costs. The task list is passed to the Parallel Simulator, which calculates the data transfer and memory read/write times caused by different parallel strategies. The simulator outputs the optimal parallel configuration, recording the top three best configurations.

GUIDE then evaluates the performance of different configurations based on inference time or throughput. For each configuration, the simulation results give the best parallel configuration and the corresponding inference time. And the final throughput is derived as follows:
\[
\text{single\_gpu\_throughput} = \frac{1}{\text{TPOT}}
\]
and
\[
\text{multi\_gpu\_throughput} = \frac{N \cdot T_{\text{single}}}{1 + \log_2(P_{\text{TP}})},
\]
where \(\text{TPOT}\) represents the token processing time per operation for a single GPU, \(N\) is the number of GPUs, \(T_{\text{single}}\) denotes the single-GPU throughput, and \(P_{\text{TP}}\) stands for the parameter parallelism.

TTFT for the prefill phase is simply the time per token for the prefill phase.

These results are used to identify the top three configurations with the best performance based on inference time or throughput, where the configuration with the lowest inference time and highest throughput is considered optimal.

\subsection{Model Analyzer}
\label{sec:technique2}

\begin{figure}[htbp]
    \centering
    \includegraphics[width=0.47\textwidth]{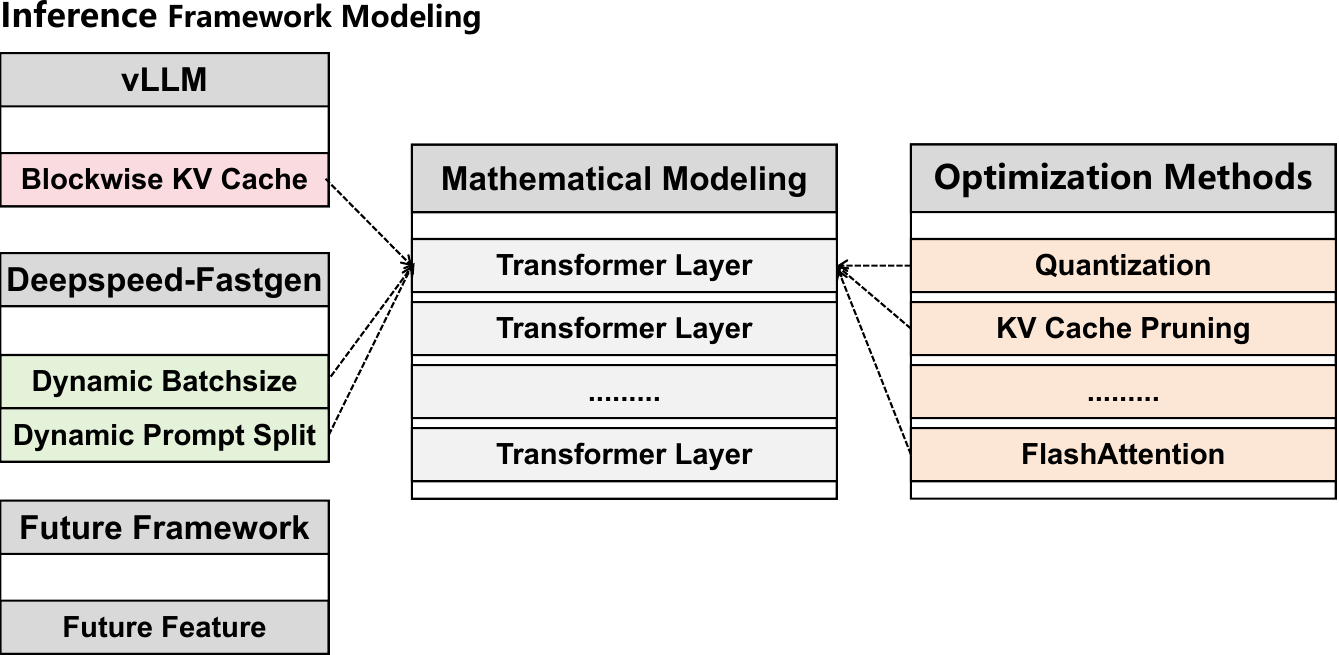} 
    \caption{Overview of the Model Analyzer technique in GUIDE. The figure shows how inference framework modeling and optimization methods collaboratively act on each transformer layer to enable more accurate modeling and evaluation of LLM inference.}
    \label{fig:model_analyzer_overview}
\end{figure}

In this section, we describe the second key component of GUIDE, which is the Model Analyzer. This technique focuses on optimizing memory utilization and computational efficiency by dynamically adjusting key parameters, such as batch size, sequence length, and KV cache size, based on the available GPU memory and model requirements. As shown in Figure~\ref{fig:model_analyzer_overview}, the Model Analyzer integrates dynamic batch size adjustment, sequence length adjustment, and key-value (KV) cache optimization with H2O to simulate throughput while using these optimization and inference framework.

\subsubsection{Dynamic Batch Size Adjustment}

The dynamic batch size adjustment mechanism is designed to optimize the batch size while ensuring it adheres to the constraints imposed by GPU memory limitations. This approach consists of two primary steps: calculating the memory required per request and dynamically adjusting the batch size.

First, the memory required for the key-value (KV) storage per request is calculated as:
\[
\text{kv\_byte\_per\_request} = 2 \times N \times H \times S \times B \times kv,
\]
where:
- \( N \) represents the number of hidden layers,
- \( H \) is the hidden size of the model,
- \( S \) is the sequence length,
- \( B \) is the batch size,
- \( kv \) denotes the memory in bytes required for each key-value pair.

To isolate the memory consumption independent of the batch size, the memory required for KV storage per request without the influence of batch size is expressed as:
\[
\text{kv\_byte\_per\_request\_without\_batchsize} = 2 \times N \times H \times S \times kv.
\]

Given the GPU's total memory capacity \( C_{\text{gpu}} \) and the memory occupied by the model \( C_{\text{model}} \), the available memory for KV storage is:
\[
C_{\text{available}} = C_{\text{gpu}} - C_{\text{model}}.
\]
The maximum allowable batch size under these memory constraints is determined as:
\[
B_{\text{max}} = \left\lfloor \frac{C_{\text{available}}}{\text{kv\_byte\_per\_request\_without\_batchsize}} \right\rfloor.
\]

If the input batch size \( B_{\text{input}} \) exceeds this maximum allowable batch size, the batch size is adjusted to:
\[
B = \min(B_{\text{input}}, B_{\text{max}}).
\]

To further optimize memory usage and enhance prompt processing performance, the adjusted batch size is divided into smaller sub-batches. The sub-batch splitting process is designed to ensure that each sub-batch fits within the available memory while maintaining computational efficiency.

The size of each sub-batch is determined by a dynamically chosen parameter, denoted as \( \text{split\_size} \), which balances memory constraints with the need for efficient computation. Specifically, the batch size \( B \) is divided into sub-batches of size \( \text{split\_size} \), ensuring that the number of sub-batches is maximized without exceeding memory limitations. The number of sub-batches is determined by the ratio of the batch size to the split size, with any remaining portion of the batch forming an additional sub-batch.

\subsubsection{Dynamic Sequence Length Adjustment}
The core of the dynamic sequence length adjustment is to optimize the sequence length based on the GPU's available memory, ensuring that memory constraints are respected while maximizing efficiency. The process consists of two main steps: calculating the memory required per request and adjusting the sequence length dynamically.

First, the memory required for each request, without considering the sequence length, is calculated as:

\[
\text{kv\_byte\_per\_request\_without\_seqlen} = 2 \times N \times H \times B \times kv,
\]

Given the available memory, the maximum sequence length \( S_{\text{max}} \) that can fit within the memory constraints is calculated as:

\[
S_{\text{max}} = \left\lfloor \frac{C_{\text{available}}}{\text{kv\_byte\_per\_request\_without\_seqlen}} \right\rfloor.
\]

If the input sequence length exceeds the maximum sequence length allowed by the available GPU memory, the sequence length is adjusted to fit within the memory constraints.

Additionally, to simulate the impact of long prompts, the sequence is divided into smaller segments. The size of each segment is determined by a base value, which is then multiplied by a tuning factor to allow for fine-tuning. The total number of splits required is calculated by dividing the input sequence length by the split size and rounding up to the nearest integer.

Finally, the sequence length is adjusted based on the number of splits, with the adjusted sequence length being the product of the number of splits and the split size. If the adjusted sequence length exceeds the original sequence length, the adjustment is applied to align the sequence length with the available memory. Otherwise, the sequence length remains unchanged.

The final adjusted sequence length is returned, which will either be the adjusted length if it does not exceed the original input length, or the original sequence length if no adjustment was necessary.

\subsection{Parallel Simulator}
\label{sec:technique3}

In this section, we introduce the third key technique of GUIDE: the Parallel Simulator. This simulator focuses on optimizing task execution in multi-GPU environments by simulating hybrid parallelism, combining Data Parallelism (DP) and Tensor Parallelism (TP). The goal is to find the optimal parallel configuration that minimizes the total execution time of the tasks.

\begin{figure}[htbp]
    \centering
    \includegraphics[width=0.47\textwidth]{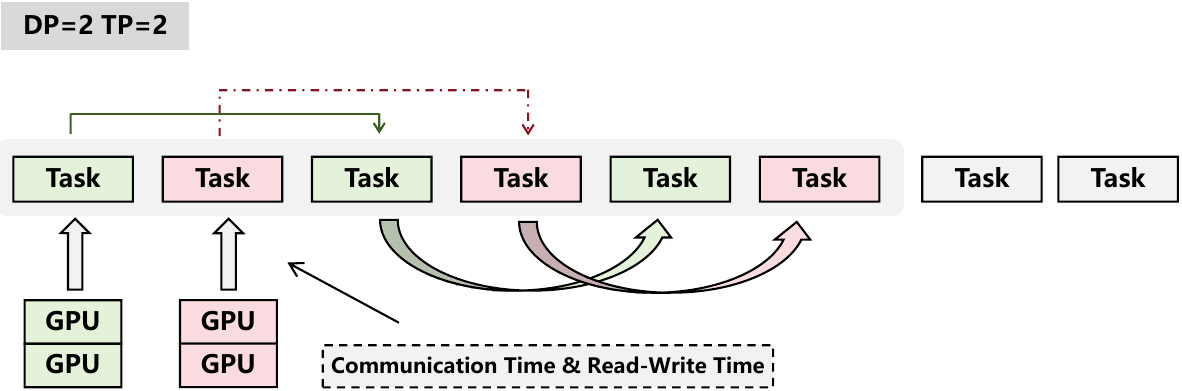} 
    \caption{Parallel execution flow in the Parallel Simulator. The figure shows how tasks are distributed using DP and TP, with communication and read-write overheads modeled.}
    \label{fig:parallel_simulator_process}
\end{figure}

As shown in Figure~\ref{fig:parallel_simulator_process}, the Parallel Simulator handles the task execution flow by simulating both Data Parallelism and Tensor Parallelism to minimize the execution time across multiple GPUs.

\subsubsection{Task and GPU Models}
The core of the simulator consists of two main models: the \textit{Task} class and the \textit{GPU} class.

- Task Model: Each task in the simulator is represented by a \texttt{Task} object, which holds the data size and the compute load of the task.
- GPU Model: The \texttt{GPU} class captures the hardware characteristics of a single GPU, including its compute performance, memory bandwidth, communication bandwidth, and latency.

\subsubsection{Roofline Model for Performance Analysis}
To analyze whether a task's performance is constrained by compute power or memory bandwidth, the simulator uses the Roofline model from LLM-Viewer. This model plots the performance ceiling of a task based on the available compute and memory resources. Specifically, the Roofline model can be expressed as:

\[
\text{Performance} = \min \left( \frac{\text{compute\_load}}{\text{compute\_time}}, \frac{\text{memory\_bandwidth}}{\text{data\_size}} \right)
\]
Where:
- \( \text{compute\_load} \) is the number of floating point operations (FLOPs) required by the task.
- \( \text{compute\_time} \) is the time taken by the task on the GPU based on its compute performance (TFLOPS).
- \( \text{memory\_bandwidth} \) is the rate at which data can be transferred between memory and the processor.

The Roofline model helps to identify the bottleneck for each task—whether the task is memory-bound or compute-bound—and informs the simulation of execution time.

\subsection{Task Execution and Hybrid Parallelism Modeling}

\subsubsection{Time Calculation}

The task execution time can be broken down into three key components. The first component is the \textit{compute time}, which is determined by the total computation load of the task divided by the computational power of the GPU. Mathematically, it is expressed as:

\[
\text{compute\_time} = \frac{\text{compute\_load}}{\text{gpu\_compute\_power}}
\]

The second component is the \textit{data read/write overhead}, which refers to the time required to read and write data to and from the GPU memory. This is calculated by dividing the total data size by the GPU memory read/write bandwidth:

\[
\text{data\_read\_write\_time} = \frac{\text{total\_data}}{\text{gpu\_memory\_bw}}
\]

The third component is the \textit{data transfer overhead}, which represents the time required to transfer data between system memory and the GPU memory. This is determined by the total amount of data transferred and the GPU memory communication bandwidth:

\[
\text{data\_transfer\_time} = \frac{\text{total\_data\_transferred}}{\text{gpu\_memory\_comm\_bw}}
\]

The total task execution time is then the sum of these three components.

\subsubsection{Hybrid Parallelism (DP and TP) Modeling}

Hybrid parallelism, which combines Data Parallelism (DP) and Tensor Parallelism (TP), plays a crucial role in optimizing task execution. In the context of hybrid parallelism, the task is divided into multiple groups, with each group of tasks processed in parallel across several GPUs. In Data Parallelism (DP), the task is split into batches, with each batch processed by multiple GPUs in parallel. The execution time for each batch is determined by the slowest GPU in that batch, taking into account any communication delays between GPUs.

On the other hand, Tensor Parallelism (TP) involves splitting the model parameters themselves across multiple GPUs. Within each batch, each GPU processes a portion of the model's parameters, and all GPUs in the group work simultaneously on a given task. This strategy helps in reducing the computational load on each individual GPU and improves overall performance by leveraging the parallelism of the model's structure.

The hybrid parallelism approach combines these two techniques, enabling the simulation of a system where tasks are distributed across multiple GPUs with both data and tensor parallelism. This results in a more efficient utilization of the available resources and a reduction in the total execution time for large-scale tasks.

\subsubsection{Simulating Hybrid Parallelism}

A key feature of the simulator is its ability to simulate hybrid parallelism, which combines both Data Parallelism (DP) and Tensor Parallelism (TP). The \texttt{simulate\_task\_execution} function enables the simultaneous use of both parallel strategies. The function first generates all possible combinations of DP and TP configurations using the \texttt{get\_configurations} method, ensuring that the number of GPUs used does not exceed the available resources.

The simulation process begins with configuration generation, where the \texttt{get\_configurations} method creates all feasible DP and TP configurations by varying the number of GPUs allocated to the DP and TP tasks. The objective here is to explore the full range of possible configurations and identify the optimal balance between the two parallel strategies. 

Once the configurations are generated, the simulator proceeds with the task execution simulation. The \texttt{simulate\_hybrid\_parallel} method is used to simulate the execution of tasks under each DP and TP configuration. This method considers important factors such as compute time, memory bandwidth, and communication delays. By accounting for these variables, the simulator provides a detailed assessment of how the tasks are executed in a hybrid parallel environment.

Finally, after simulating the execution for all configurations, the simulator evaluates the total execution time for each configuration. The optimal configuration is the one that minimizes the total execution time across all tasks. This configuration is selected based on its ability to best balance the execution times of both DP and TP tasks, while adhering to the constraints of available GPU resources. The goal is to achieve the most efficient use of resources, ensuring that the task execution is optimized across the hybrid parallel environment.

\section{Implementation}

In this section, we describe the implementation of our system, which is built using Python and consists of a frontend and a backend. The frontend serves as the user interface, while the backend handles core functionalities, including user request processing and performance analysis.

\subsection{Backend Implementation}

The backend, implemented in Python, acts as the core component for processing user requests and executing logic. It is composed of three main modules: `inference engine`, `model analyzer`, and `parallel simulator`. The `inference engine` module serves as the interface, accepting user inputs such as budget, selected model, generated sequence length, and input sequence length. Based on these inputs, it triggers backend processes to provide outputs, including the hardware configuration with the highest throughput, the configuration with the minimum inference time, and the top three hybrid parallelism configurations optimized for performance and budget constraints. To achieve these outputs, `inference engine` coordinates with `model analyzer` and `parallel simulator`, calculates performance metrics, and presents the results to the user for informed decision-making.

The second module, `model analyzer`, built on the `llm-viewer` framework, evaluates the performance of inference frameworks and optimizations. This module integrates frameworks such as \texttt{vllm} and \texttt{fastgen} for model assessment and considers key-value (KV) store optimizations, including \texttt{h2o}, to enhance data handling during inference. By analyzing the impact of these frameworks and optimizations on throughput and inference time, `model analyzer` identifies the most suitable configurations for the user's specific requirements.

The third module is a hybrid parallelism simulator designed to test various parallelization strategies within the constraints of the user's budget. It evaluates combinations of data parallelism, pipeline parallelism, and model parallelism to determine configurations that minimize inference time and maximize throughput. By considering the effect of budget constraints on these strategies, `parallel simulator` identifies optimal configurations that balance performance with resource availability, enabling users to achieve efficient model deployment under limited resources.

\subsection{User Interface}

\begin{figure}[htbp]
    \centering
    \includegraphics[scale=0.32]{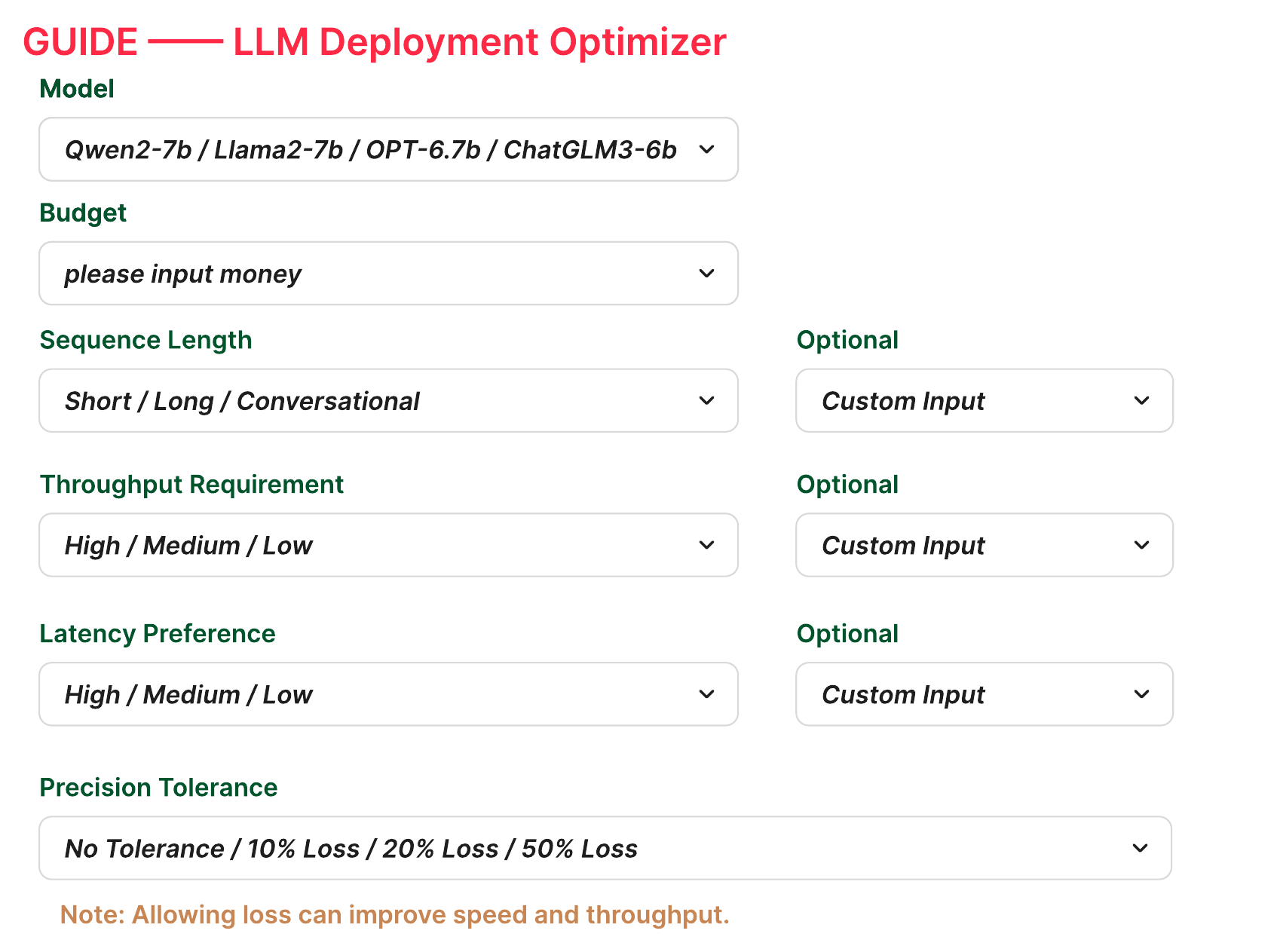} 
    \caption{UI layout. The figure shows the user interface of GUIDE, where users can specify model, budget, sequence length, throughput, latency, and precision preferences, along with optional custom inputs.}
    \label{fig:ui}
\end{figure}

A web-based user interface (UI) was developed to configure simulation parameters and display results. The UI allows users to select a model from a dropdown menu, input a budget, and configure parameters such as sequence length, throughput requirement, latency preference, and precision tolerance. Among these, sequence length, throughput requirement, and latency preference provide both predefined options for users unfamiliar with the parameters and custom input fields for users requiring precise control. Users can initiate the computation process, and the results, including relevant performance metrics, are displayed in an organized format. The interface layout is shown in Figure~\ref{fig:ui}.

\section{Evaluation}

\subsection{Experimental Setup}
We evaluate the accuracy and performance of our simulator by comparing simulated results with real experimental data under various configurations. The primary goal of the evaluation is to model and analyze the behavior of large language model (LLM) inference tasks in real-world environments, then compare it with the predictions of our simulator. The evaluation involves two parallel tracks: real-world execution on actual hardware and simulated execution within the modeling framework, with the results from both tracks used to compute prediction errors as key evaluation metrics.

In real-world experiments, tasks are designed to simulate common LLM inference scenarios. Each task involves a set of input prompts, where the prompt lengths are randomly generated with fixed values to represent realistic usage patterns. For the outputs, the simulated tasks generate fixed-length sequences, while in real-world tasks, we intentionally limit the maximum output length to align with the fixed output length in the simulation. This approach introduces some unavoidable discrepancies, as real-world model outputs are inherently variable and difficult to fully control.

Each model was tested with varying configurations to evaluate the simulator’s accuracy under diverse scenarios. The configurations included batch sizes of {4, 8, 16, 32}, which represent different levels of parallel workloads, and prompt lengths of {128, 256, 512, 1024}, which reflect diverse input complexities. The output length was fixed at 256 tokens in the simulation environment for standardization, while in real-world experiments, the maximum output length was constrained to 256 tokens to align with the simulation. These configurations were chosen to comprehensively cover real-world usage patterns and provide a robust evaluation of the simulator’s performance.

\begin{figure*}[b]
    \centering
    \includegraphics[width=\textwidth]{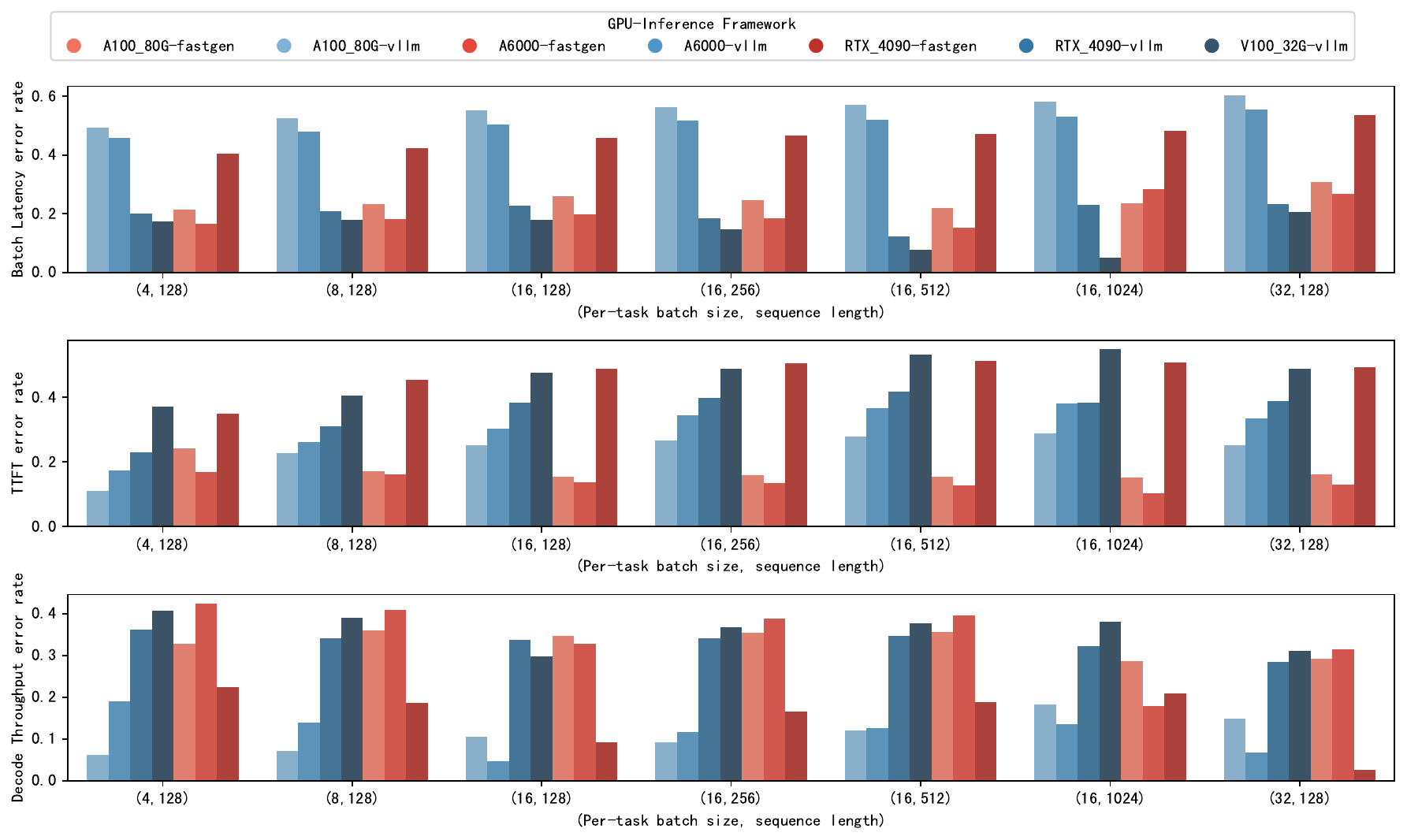} 
    \caption{Comparison of performance metrics on a single GPU between vLLM and FastGen across different input lengths and batch sizes. The figure illustrates the batch latency, TTFT, and decode throughput for both models under identical conditions.}
    \label{fig:single}
\end{figure*}

In the simulator, the same configurations, including hardware settings, batch sizes, prompt lengths, and model architectures, are replicated. The simulator generates predictions for three key metrics: Time-to-First-Token (TTFT), which measures the time required to produce the first token; Decode Throughput, defined as the throughput during the decoding phase, measured in tokens processed per second; and Batch Latency, representing the total time required to complete the processing of an entire batch. These predictions are compared against real-world measurements to calculate the error values, which serve as indicators of the simulator’s accuracy.

The experiments are conducted on diverse hardware, software, and model configurations. For single-GPU experiments, we tested vLLM on NVIDIA A100 80G, V100 32G, A6000, and RTX 4090 GPUs, while FastGen was evaluated on NVIDIA A100 80G, A6000, and RTX 4090. For multi-GPU experiments, we used NVIDIA A100 80G, A6000, and RTX 4090 under tensor parallelism for both frameworks. The large language models assessed include THUDM/chatglm3-6b, Qwen/Qwen2-7B, facebook/opt-6.7b, and meta-llama/Llama-2-7b-hf.

\subsection{Results Analysis}
\subsubsection{Single-GPU Evaluation}
In our single-GPU experiments, we evaluated a variety of models to assess their error performance across different configurations. To simplify the data presentation, we report the mean error rates across all tested models, as summarized in Table~\ref{tab:error_metrics}. Specifically, the mean errors for Batch Latency, TTFT, and Decode Throughput are 33.04\%, 9.9\%, and 26.80\%, respectively, for \texttt{vLLM}, and 32.74\%, 12.30\%, and 30.10\% for \texttt{FastGen}. Figure~\ref{fig:single} provides a detailed view of these errors across different GPU types, input lengths, and batch sizes. This aggregated presentation offers insight into the general trends and variations in prediction errors under diverse configurations.

\begin{table}[ht]
\centering
\caption{Comparison of Error Metrics on a Single GPU.}
\vspace{0.5em} 
\label{tab:error_metrics}
\begin{tabular}{|l|c|c|}
\hline
\textbf{Error Metric}      & \textbf{vLLM (\%)} & \textbf{FastGen (\%)} \\ \hline
\textbf{Batch Latency}     & 33.04              & 32.74                 \\ \hline
\textbf{TTFT}              & 9.9              & 12.30                 \\ \hline
\textbf{Decode Throughput} & 26.80              & 30.10                 \\ \hline
\end{tabular}
\end{table}


\begin{figure*}[b]
    \centering
    \includegraphics[width=0.9\textwidth]{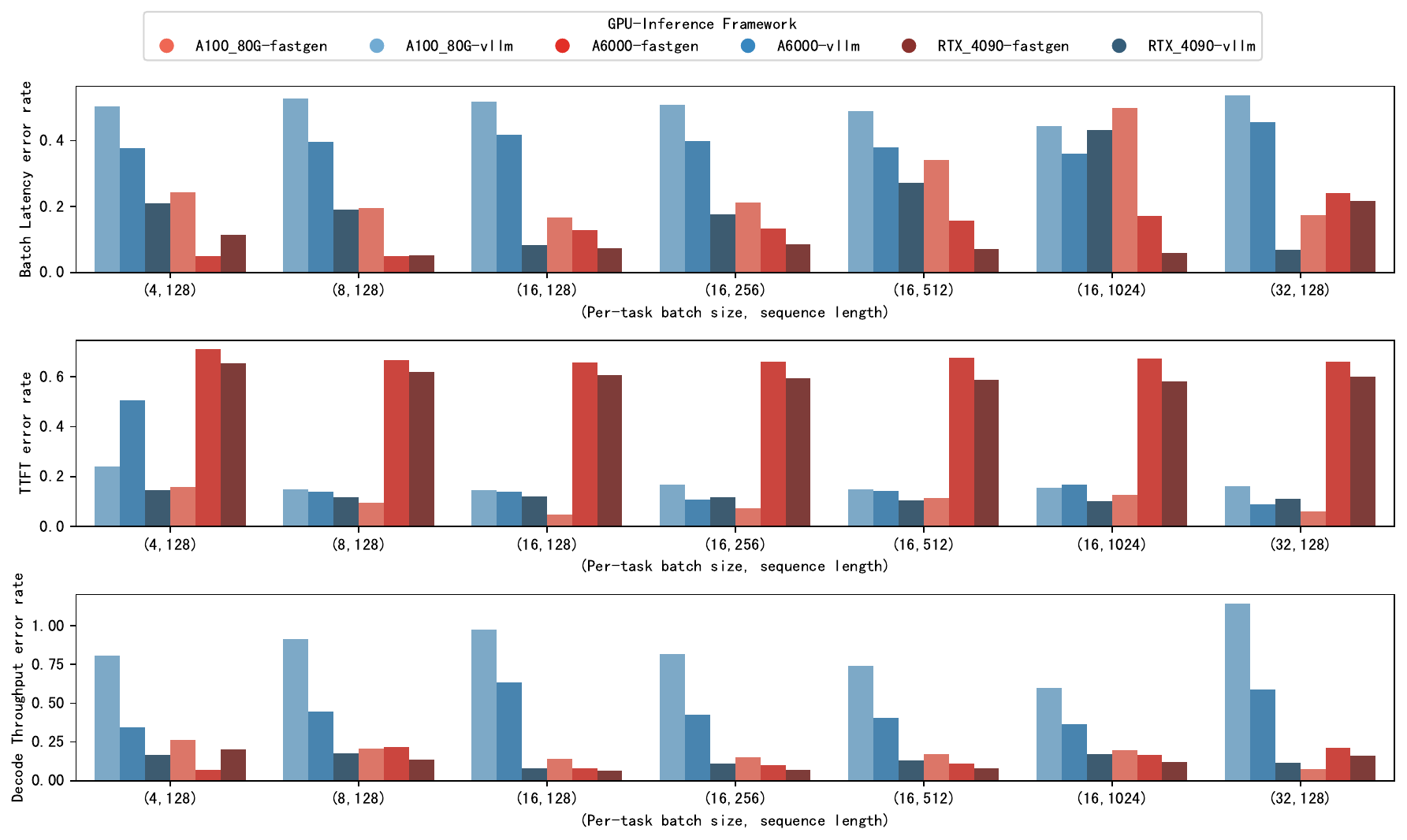} 
    \caption{Comparison of performance metrics on multiple GPUs between VLLM and FastGen across different input lengths and batch sizes. The figure illustrates the batch latency, TTFT, and decode throughput for both models under identical conditions in a multi-GPU setup.}
    \label{fig:multi}
\end{figure*}

The results presented in Table~\ref{tab:error_metrics} and Figure~\ref{fig:single} reveal the complexities in accurately predicting GPU performance across diverse configurations. While the overall trends show increasing errors with larger input lengths and batch sizes, the \texttt{nvidia\_A100\_80G} exhibits relatively higher errors compared to other GPUs. This phenomenon can be attributed to the unique architecture and scaling characteristics of the A100. Specifically, the high computational density and parallelism of the A100 may lead to greater sensitivity to workload imbalance or suboptimal utilization of its hardware features. For example, under extreme configurations such as very large batch sizes or input lengths, the model's assumptions about uniform resource usage may deviate significantly from the actual hardware behavior.

Additionally, unmodeled factors such as kernel-level behavior, synchronization overhead, and cache utilization can further exacerbate the discrepancies between predicted and actual performance. These issues are particularly pronounced in high-performance GPUs like the A100, where the complexity of the hardware amplifies even minor inefficiencies in workload execution. Overall, these findings highlight the inherent challenges of building precise performance models, especially for cutting-edge GPUs operating under diverse and demanding workloads.

\subsubsection{Multi-GPU Evaluation}
In our multi-GPU experiments, we evaluated the error performance of \texttt{vLLM} and \texttt{FastGen} across different configurations. To simplify the presentation, Table~\ref{tab:error_metrics} reports the averaged error rates for each model, focusing on three core metrics: Batch Latency, TTFT, and Decode Throughput. For \texttt{vLLM}, the errors are 25.19\%, 30.40\%, and 16.00\%, respectively, while \texttt{FastGen} shows errors of 28.06\%, 42.30\%, and 23.90\%. A detailed breakdown of these errors is provided in Figure~\ref{fig:single}, which visualizes the trends across different GPU types, input lengths, and batch sizes. The results highlight considerable variability in errors under diverse conditions, offering insights into the performance trade-offs involved in multi-GPU setups.

\begin{table}[ht]
\centering
\caption{Comparison of Error Metrics Across Multiple GPUs.}
\vspace{0.5em} 
\label{tab:error_metrics}
\begin{tabular}{|l|c|c|}
\hline
\textbf{Error Metric}      & \textbf{vLLM (\%)} & \textbf{FastGen (\%)} \\ \hline
\textbf{Batch Latency}     & 25.19              & 28.06                 \\ \hline
\textbf{TTFT}              & 30.40              & 42.30                 \\ \hline
\textbf{Decode Throughput} & 16.00              & 23.90                 \\ \hline
\end{tabular}
\end{table}

The multi-GPU results in Table~\ref{tab:error_metrics} and Figure~\ref{fig:single} reveal several key factors influencing prediction errors. Firstly, errors generally decrease compared to single-GPU setups, reflecting better workload distribution and parallelism. However, as input lengths and batch sizes increase, errors begin to rise, driven by more complex inter-GPU communication and synchronization overheads. These overheads, while partially mitigated by modern GPUs like the \texttt{nvidia\_A100\_80G}, still contribute to deviations from predicted performance due to variable latency in data transfer and kernel synchronization.

In addition, the higher Decode Throughput errors observed for \texttt{FastGen} suggest greater sensitivity to scaling inefficiencies in multi-GPU environments. This could be linked to workload partitioning strategies, where imbalances in processing across GPUs lead to underutilization of computational resources or bottlenecks in specific GPUs. Furthermore, unmodeled system-level factors, such as memory contention and cache behavior, exacerbate these discrepancies, especially under extreme configurations.



 
\section{limitations \& Future Work}

One of the main limitations of this study lies in the simulation accuracy. The models used for performance evaluation do not fully account for real-world conditions such as CPU-GPU interactions, power supply fluctuations, thermal dynamics, and the impact of other concurrently running applications. These factors, which significantly influence performance in practical scenarios, are difficult to simulate accurately and are not included in our analysis. As a result, the performance results may differ from what would be observed in a real-world environment, and the conclusions drawn from these simulations may not perfectly represent actual performance under varied operational conditions.

Another limitation stems from the abstraction used in modeling the inference framework. While the framework captures many important aspects of task scheduling and resource allocation, it is not a comprehensive representation of every component and interaction involved in a real inference system. Key factors such as hardware-level optimizations, architectural differences between GPUs, and the dynamic behavior of various software layers are simplified or omitted. This partial modeling inevitably introduces errors and discrepancies between the simulated and real-world performance, particularly in complex, heterogeneous environments.
 
\section{Related Work}

\subsection{Mathematical Modeling and Simulator-Based Performance Evaluation}

Mathematical modeling and simulation frameworks are crucial for optimizing large language model (LLM) inference systems. Recent advancements focus on system-level efficiency and user-centric performance metrics. Andes\cite{liu2024andes} improves Quality-of-Experience (QoE) in text streaming by dynamically allocating GPU resources, achieving up to 3.2× QoE gains under high loads. Etalon critiques traditional metrics like throughput and latency, introducing the fluidity index to better capture real-time performance.

Simulation tools such as Vidur \cite{agrawal2024vidur} and GenZ \cite{bambhaniya2024demystifying} provide valuable insights into platform requirements. Vidur combines experimental profiling with predictive modeling, maintaining less than 9

Optimization techniques also improve resource utilization. Nanoflow \cite{zhu2024nanoflow} uses nano-batching and operation-level pipelines to overlap memory, compute, and network operations, achieving up to 1.91× throughput improvements. Uellm \cite{he2024uellm} combines resource profiling, batch scheduling, and dynamic deployment strategies to reduce latency and enhance GPU utilization, demonstrating up to 4.98× throughput gains.

Despite these advancements, many tools focus on specific aspects or deployment scenarios. Vidur and Etalon primarily address static profiling and metric refinement, which may overlook dynamic, real-world workloads. Metrics like "idealized runtime" enable cross-model comparisons but abstract away critical workload variability and system contention factors. These limitations highlight the need for comprehensive frameworks that integrate user-centric metrics, real-time adaptability, and cost-aware optimization strategies.

\subsection{Optimization Strategies for Large-Scale Models}

Optimizing inference for large-scale language models (LLMs) focuses on memory efficiency, throughput, and latency. Several strategies target different parts of the inference pipeline.

Memory management is crucial for long-context generation. Infinigen \cite{lee2024infinigen} and H2O \cite{zhang2023h2o} improve Key-Value (KV) cache management. Infinigen speculates essential KV entries, reducing fetch overhead and improving offloading-based systems by up to 3×. H2O treats KV eviction dynamically, improving throughput by up to 29×. However, both depend on specific hardware optimizations, limiting generalizability.

Algorithmic innovations like vllm’s PagedAttention and Razorattention \cite{tang2024razorattention} boost memory efficiency. Vllm reduces KV cache waste using virtual memory, achieving 2–4× throughput improvements for longer sequences. Razorattention compresses the cache by over 70

For computational efficiency, Deepspeed-fastgen and vllm offer significant gains. Deepspeed-fastgen improves throughput by 2.3× and reduces tail latency with dynamic prompt generation. Vllm’s IO-aware tiling algorithm cuts memory reads/writes, achieving 3× speedup for GPT-2 \cite{radford2019language} on long sequences.

Sarathi-Serve \cite{agrawal2024taming} optimizes throughput and latency by refining the prefill and decode phases. Its chunked-prefills and stall-free scheduling improve serving capacity by up to 5.6× for models like Falcon-180B \cite{almazrouei2023falcon}, making it suitable for real-time applications.

\subsection{Pre-trained Models for Large-Scale Inference}

Pre-trained large language models (LLMs) are crucial in AI, with advancements in scalability, accessibility, and fine-tuning. The Qwen2 series, ranging from 0.5 to 72 billion parameters, excels in benchmarks like CMMLU \cite{hendrycks2020measuring} and HumanEval \cite{chen2021evaluating}, supports 30 languages, and promotes community-driven fine-tuning. The Llama2 series, with models like llama2-Chat, performs well in dialogue tasks, offering an open alternative to proprietary models.

The chatglm family, especially GLM-4, competes with proprietary models like GPT-4 \cite{achiam2023gpt}, adding tool integration for complex tasks, which emphasizes alignment and multimodal integration, especially for Chinese applications. The OPT initiative, with models from 125M to 175B parameters, focuses on reproducibility and sustainability, achieving GPT-3-level performance while reducing carbon footprints.

Despite these advancements, challenges remain in balancing scalability, efficiency, and alignment. Models like Qwen2 and GLM-4 require substantial computational resources, limiting access for smaller organizations. Improving alignment across languages and cultures is an ongoing research area, with future work focusing on resource-efficient training and stronger alignment techniques to increase accessibility and societal impact.
 
\section{Conclusion }

In conclusion, GUIDE provides a robust and scalable framework for optimizing the inference performance of large-scale language models. By systematically addressing critical bottlenecks identified through experimentation, such as memory inefficiencies, latency divergence, and multi-GPU scaling issues, GUIDE bridges the gap between theoretical model performance and practical deployment requirements. Its integration of dynamic modeling, simulation-based optimization, and intelligent deployment strategies equips researchers and practitioners with practical tools to navigate the complexities of real-world environments. With further advancements in its capabilities and enhanced predictive accuracy, GUIDE has the potential to become an indispensable resource for deploying large-scale models efficiently, cost-effectively, and adaptively across heterogeneous environments.

\bibliographystyle{unsrt}
\bibliography{reference}

\end{document}